\documentclass{INTERSPEECH2023}
\usepackage[utf8]{inputenc}
\usepackage{graphicx}
\usepackage{cite,multirow,balance}
\usepackage{amsmath}
\usepackage{float}

\interspeechcameraready

\title{SALTTS: Leveraging Self-Supervised Speech Representations for \newline improved Text-to-Speech Synthesis}
\name{Ramanan Sivaguru, Vasista Sai Lodagala, S Umesh}
\address{Speech Lab, Indian Institute of Technology Madras, India}
\email{reachramzz25@gmail.com, vasista.lodagala@gmail.com, umeshs@ee.iitm.ac.in}

\begin{document}

\maketitle
 
\begin{abstract}


While FastSpeech2 aims to integrate aspects of speech such as pitch, energy, and duration as conditional inputs, it still leaves scope for richer representations. As a part of this work, we leverage representations from various Self-Supervised Learning (SSL) models to enhance the quality of the synthesized speech.
In particular, we pass the FastSpeech2 encoder's length-regulated outputs through a series of encoder layers with the objective of reconstructing the SSL representations. In the SALTTS-parallel implementation, the representations from this second encoder are used for an auxiliary reconstruction loss with the SSL features. The SALTTS-cascade implementation, however, passes these representations through the decoder in addition to having the reconstruction loss.
The richness of speech characteristics from the SSL features reflects in the output speech quality, with the objective and subjective evaluation measures of the proposed approach outperforming the baseline FastSpeech2.

\end{abstract}

\vspace{-0.5em}
\section{Introduction}
\vspace{-0.5em}
\begin{figure*}[t]
\centering
\includegraphics[width=\textwidth]{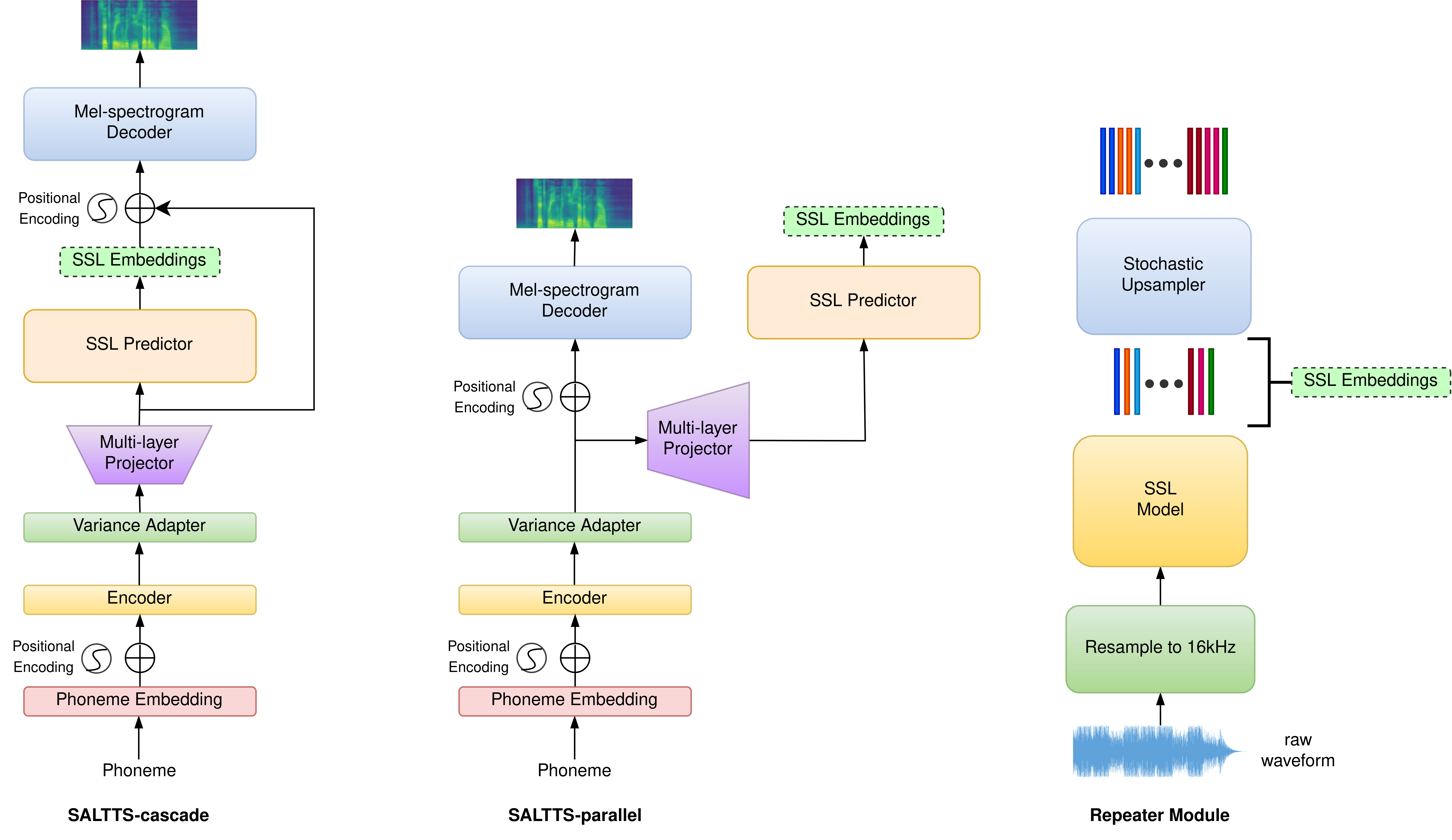}
\caption{Illustration of the parallel and cascade variants of the SALTTS architecture.}
\label{fig:saltts arch}
\end{figure*}

Over the past few years, text-to-speech (TTS) technology has advanced remarkably, revolutionizing how people communicate with machines. Today, TTS finds applications ranging from audiobooks to virtual assistants. The rapid expansion of the internet and social media platforms is leading to a surge in the need for TTS systems that can produce high-quality and natural-sounding speech.

Building a high-quality Text-to-Speech (TTS) system requires a large amount of labeled data, which can be expensive and time-consuming. Self-supervised Learning (SSL) for speech related tasks has emerged as a promising approach in the recent years, and addresses such labeled data constraints. 
wavlm \cite{wavlm}, data2vec \cite{data2vec_norm}, HuBERT \cite{hsu2021hubert} and wav2vec 2.0 \cite{wav2vec} are few of the SSL paradigms that have shown remarkable performance over various downstream tasks such as speech recognition, speaker identification, and emotion recognition \cite{superb}. 
However, the application of SSL representations within the TTS setup has been relatively under-studied.
This work aims to investigate if the insights gained by the SSL models can be utilized to improve the synthesized speech quality in TTS systems.

This paper uses FastSpeech2 \cite{fs2} model as a baseline system for conducting the experiments. However, the ideas proposed in this paper are equally applicable to the JETS (Jointly Training FastSpeech2 and HiFi-GAN for End to End Text to Speech) \cite{jets} and VITS (Variational Inference with adversarial learning for end-to-end Text-to-Speech) \cite{vits} framework. The FastSpeech2 model is a strong baseline system since it has also been trained to predict pitch and energy information of the audio from the text transcription in addition to the mel-spectrogram. This additional feature helps the model capture various speech variations and nuances. Furthermore, the FastSpeech2 model's non-auto regressive nature offers a distinct advantage over its auto-regressive counterparts. Faster model training and quicker inference times have been among the most desirable aspects of every deep learning algorithm, and FastSpeech2 achieves both these objectives owing to its non-auto regressive design choice.

Moreover, the model's architecture is fairly simple to comprehend, making it easier to introduce changes and updates. This characteristic is particularly beneficial, allowing us to tweak the model's structure to suit our needs and objectives better. To generate audio waveforms from mel-spectrograms, we have utilized a well-trained vocoder model based on Generative Adversarial Networks (GANs) \cite{gans}. The HiFi-GAN \cite{hifi_gan} vocoder has been the choice of the vocoder model for our experiments as it is being widely used and promising in terms of performance.
By utilizing a pre-trained HiFi-GAN model, we avoid the time-consuming and computationally expensive process of training a vocoder model from scratch.

The motivation behind this study is to explore novel ways to leverage SSL representations with improving the quality and variability of synthesized speech as the objective. To this end, we present a novel text-to-speech (TTS) model, which we refer to as SALTTS. SALTTS stands for \textbf{S}elf-supervised representations for \textbf{A}uxiliary \textbf{L}oss in \textbf{TTS} and is based on the FastSpeech2 architecture. Nevertheless, as mentioned earlier, this approach equally applies to frameworks like JETS, VITS, etc. We develop two variants for this newly proposed model namely, SALTTS-parallel and SALTTS-cascade. The primary objective of these architectures is to introduce an additional encoder block (SSL predictor) that is capable of predicting the embeddings generated by the SSL models. By doing so, we aim to enhance the capability of the model to capture a broader range of speech variations in addition to the energy and pitch information that FastSpeech2 has incorporated. The addition of the SSL predictor block enables the model to learn more nuanced representations of speech, which could eventually translate into better quality of synthesized speech. Overall, we believe that our proposed methods have the potential to improve over the state-of-the-art in speech synthesis significantly.

Our paper aims to advance speech synthesis by exploring the intersection between text-to-speech (TTS) synthesis and self-supervised learning (SSL). In particular, our focus is on identifying novel research directions and opportunities for leveraging the rich SSL representations into improving the performance and efficiency of TTS models. The major contributions from this work are as follows:
\begin{itemize}
    \item We develop SALTTS-parallel and SALTTS-cascade, both of which aim to introduce information from SSL representations into the FastSpeech2 architecture.
    \item During inference, SALTTS-parallel operates exactly as FastSpeech2. This helps the model learn from the additional information presented by the SSL representations during training, while retaining the inference speeds that FastSpeech2 promises.
    \item An additional repeater module has been developed to account for the different sampling rates, window sizes and hop-lengths that FastSpeech2 and the SSL models operate at. This module is quintessential to our proposed approach as it ensures better alignment between the predicted and ground truth SSL embeddings.
\end{itemize}


\vspace{-0.5em}
\section{Related Work}
\vspace{-0.5em}

The intersection of Self Supervised Learning and Text-to-Speech synthesis is an area that has received limited attention within the speech research community. Prior studies have largely focused on predicting mel-spectrograms as a means of generating speech \cite{natural_speech} \cite{glow_tts}. However, a new approach has emerged wherein discrete representation units, such as clusters of SSL embeddings, are predicted instead of mel-spectrograms \cite{vqtts} \cite{s2s_with_discrete_units} \cite{speech_resynthesis}. Subsequently, speech is generated from these discretized representations. To the best of our knowledge, no research has been conducted yet that leverages SSL embeddings to aid in generation of mel-spectrograms.

\vspace{-0.5em}
\section{Methodology}
\vspace{-0.5em}
In  the following subsections, we elaborate on the architecture of the proposed systems and the multi-task loss which together aim to enhance the embeddings from the acoustic model. Furthermore, we motivate the need for a repeater module over the SSL representations and describe the functionality of the same.

\vspace{-0.5em}
\subsection{Motivation}
\label{sec:motiv}
\vspace{-0.5em}

The architecture of FastSpeech2 \cite{fs2} addresses one of the key issues with FastSpeech \cite{fs} by training using the ground truth targets instead of the outputs from the teacher model. In addition, FastSpeech2 extracts duration, pitch and energy information from the waveform and uses them as conditional inputs during the training phase, thereby introducing variability to the process.

Speech signals are complex in nature having several characteristics such as emotion, intent, emphasis and tone among many more. Having a predictor in FastSpeech2 to account for each of these facets is not practical, as rich ground truth information for each of those speech aspects is not directly available. Also, such a design would continuously increase the complexity of the model in the process of accounting for each aspect of speech.

Models pre-trained using the SSL paradigm have been observed to generalize over an array of speech tasks such as emotion detection, speaker recognition and phoneme recognition, the performance of which has been documented over SUPERB \cite{superb}. Given that the representations from these speech SSL models serve multiple downstream tasks, we leverage the embeddings from these models to bring in the various aspects of speech that have not been accounted for in the design of FastSpeech2. 

\vspace{-0.5em}
\subsection{Architecture and Variants}
\label{sec:arch}
\vspace{-0.5em}

We extract the SSL representations for every utterance of the LJSpeech dataset \cite{lj_data}. It is to be noted that these representations are 768-dimensional, given that we employ the BASE variant of the various SSL models. The embeddings generated by the variance adapter of FastSpeech2 which are 384-dimensional, are passed through a multi-layer projector which converts them to 768-dimensional embeddings. 
A 4-layer encoder block (illustrated in Fig.\ref{fig:saltts arch} as SSL predictor) takes in these embeddings from the multi-layer projector and predicts the representations produced by the SSL model during the training stage. An auxiliary L1-loss is computed between the embeddings predicted by this encoder and the representations from the SSL model.

\vspace{-0.8em}
\begin{equation}
\mathcal{L}_{aux}=\sum_{i=1}^n\left|y-\hat{y}_i\right|
\end{equation}

where $\mathcal{L}_{aux}$ is the auxiliary SSL embedding loss, $y$ and $\hat{y}$ are true and predicted SSL embedding vectors respectively and $n$ is the total no. of datapoints.

\vspace{-0.5em}
\subsubsection{SALTTS-parallel}
\label{sec:parallel}
\vspace{-0.5em}

As illustrated in Fig.\ref{fig:saltts arch}, the SALTTS-parallel variant passes the representations from the variance adapter directly to the Mel-spectrogram decoder, maintaining the flow of FastSpeech2. 
The reason we address this variant as SALTTS-parallel is that, it computes an additional L1-loss after predicting the SSL representations from the multi-layer projector, while the decoder continues to work with the embeddings from the variance adapter. 
Such a design choice ensures that we leverage the richness of the SSL representations, while continuing to maintain the inference speeds of FastSpeech2. 
It is to be noted that during inference, SALTTS-parallel functions exactly as FastSpeech2 with the same number of parameters, since the auxiliary L1-loss is used only during the training stage.

\vspace{-0.5em}
\subsubsection{SALTTS-cascade}
\label{sec:cascade}
\vspace{-0.5em}

In the design of SALTTS-cascade (Fig.\ref{fig:saltts arch}), the Mel-spectrogram decoder works with the 768-dimensional representations from the SSL predictor.
A residual connection has been added from the output of the multi-layer projector to ensure faster convergence of the model. 
The additional L1-loss however, is computed between the embeddings from the SSL predictor (before the residual connection) and the representations from the SSL model. 
The decoder's attention dimension here is set to 768 to ensure compatibility with the 768-dimensional embeddings generated by the SSL predictor.
Given the flow of SALTTS-cascade, it takes relatively longer to synthesize utterances compared to SALTTS-parallel.

\vspace{-0.5em}
\subsection{Repeater Module}
\label{sec:repeater}
\vspace{-0.5em}

As mentioned in sections \ref{sec:motiv} and \ref{sec:arch}, SALTTS attempts to benefit from the representations of SSL models. 
However, before computing an additional L1-loss between the representations from the SSL predictor introduced in FastSpeech2 (Fig.\ref{fig:saltts arch}) and the representations from SSL models, we need to deal with the mismatch in sampling frequency and hop length between the FastSpeech2 and SSL models. Most of the speech SSL models operate on waveforms which have a sampling rate of 16kHz. FastSpeech2 for LJSpeech, however, operates on speech samples with a 22.05kHz sampling rate. Moreover while generating embeddings at a frame level, most SSL models operate with a window size of 25ms and a hop length of 20ms. FastSpeech2 however, functions with a window size of 45.6ms and a hop length of 11.6ms.

To account for this mismatch in the frame-level information, we need a strategy that aligns a set of SSL representations to the embeddings from the SSL predictor of FastSpeech2. 
On performing a time-level analysis over the progression of these models, we notice that the first 5 frames from the SSL model align with the first 7 frames from FastSpeech2.
To generate an aligned set of frames, we repeat the second and fourth frame of the SSL model, to match the number of frames from FastSpeech2.
Subsequently, every 18 frames from the SSL model align with 31 frames from FastSpeech2 in the time domain.
We follow a \{2, 2, 1\}-strategy to ensure an alignment in this case. That is, for every 3 frames from the SSL model, we repeat the first frame once and the second frame once after adding some Gaussian noise sampled from a standard normal distribution. We leave the third frame as is.
For the remainder set of frames, where such a strategy could increase the number of frames, we drop frames from the end (after repetition), until we find a match.

A repeater module (illustrated in Fig.\ref{fig:saltts arch}) that works on a set pattern is therefore essential to our process as it ensures alignment in the time domain between the two set of embeddings before the additional L1-loss is computed.

\begin{table*}[t]
\begin{center}
\normalsize
\setlength\tabcolsep{10pt}
\renewcommand{\arraystretch}{1.2}
\begin{tabular}{c c c c c}
\toprule
Model  & SALTTS-Type & MOS ($95\%$ CI) & MCD  & $F_{0}$ RMSE  \\
\toprule
Ground Truth & - & 4.51 ± 0.15 & - & - \\ \hline
Baseline FastSpeech2 & - & 3.65 ± 0.2 & \textbf{10.0750 ± 0.5099} & 0.2448 ± 0.0532 \\ \hline
\multirow{2}{*}{data2vec-aqc} & Parallel & 3.85 ± 0.19 & 10.1130 ± 0.5063 & 0.2441 ± 0.0494 \\
 &  Cascade & 3.51 ± 0.22 & 10.2129 ± 0.5274 & 0.2399 ± 0.0525 \\ \hline
 \multirow{2}{*}{ccc-wav2vec 2.0} &  Parallel & 3.87 ± 0.19 & 10.1364 ± 0.5118 & 0.2415 ± 0.0481 \\
 & Cascade & 3.54 ± 0.21 & 10.1670 ± 0.5154 & 0.2389 ± 0.0494 \\ \hline
 \multirow{2}{*}{HuBERT} &  Parallel & \textbf{3.95 ± 0.18} & 10.1016 ± 0.5085 & 0.2404 ± 0.0542 \\
 & Cascade & 3.58 ± 0.2 & 10.1505 ± 0.5103 & \textbf{0.2386 ± 0.0531}\\ 
 \bottomrule \\ 
\end{tabular}
\caption{Performance of the baseline and the proposed SALTTS models over objective and subjective evaluation metrics.}
\label{tab:Baseline Comparison}
\end{center}
\end{table*}

\vspace{-0.5em}
\section{Experimental Setup}
\vspace{-0.5em}

We use the LJSpeech dataset \cite{lj_data} to train and evaluate all the baseline and proposed models.
A GAN-based well trained vocoder is used for all the experiments in this study to produce final audio wav files from the predicted mel-spectrograms. The HiFi-GAN vocoder model \cite{hifi_gan} has been chosen for this purpose as it is one of the most widely used vocoders. The HiFi-GAN model used here is pre-trained on the same LJSpeech dataset for 2.5M iterations.

For all our experiments, we consider the FastSpeech2 model with a HiFi-GAN vocoder as our baseline system.
The baseline model and the proposed architectures are set up, trained and evaluated using the ESPnet-framework \cite{espnet}. While the baseline FastSpeech2 model was trained for 1000 epochs in 3 days, the proposed SALTTS-parallel and SALTTS-cascade models take 4.5 days to train for the same number of epochs, on 4 GPUs.

\vspace{-0.5em}
\subsection{SSL Representations}
\vspace{-0.5em}

As mentioned in section \ref{sec:arch}, SALTTS-parallel and SALTTS-cascade need an SSL model to extract representations from, which would then be reconstructed by our proposed network.
Given that there are multiple speech SSL models available, we make our choice based on their performance in the Speech Enhancement task of SUPERB \cite{tsai-etal-2022-superb}, which is a generative task.
To be uniform in the choice of embedding dimensions and the SSL model sizes, we use the BASE variants of the SSL models available.
The SSL models we use for our experiments are: HuBERT \cite{hsu2021hubert}, ccc-wav2vec 2.0 \cite{lodagala2022ccc} and data2vec-aqc \cite{Lodagala2022data2vecaqcSF}.

The embeddings from the layers 9, 10 and 11 are averaged and added to be used as targets for the SSL predictor. These are the layers that have been observed to be contributing the most for the recognition and semantics tasks \cite{wavlm}.





\vspace{-0.5em}
\section{Results and Analysis}
\vspace{-0.5em}

The proposed TTS models and the baseline model have been evaluated on both subjective and objective metrics. For objective measures, mel-cepstral distortion (MCD) and log-$F_{0}$ root mean square error ($F_{0}$ RMSE) were used. MCD is a measure of how different two sequences of mel cepstra are. The smaller the MCD between synthesized and natural mel cepstral sequences, the closer the synthetic speech is to reproducing the natural speech. Log-$F_{0}$ RMSE refers to the logarithm of the root-mean-square error (RMSE) of the fundamental frequency ($f_{0}$) contour predicted by a TTS system compared to the target $f_{0}$ contour of the reference speech.

As a subjective measure we use the Mean Opinion Score (MOS). MOS is a widely used metric in TTS systems to evaluate human listeners' perceived quality of synthesized speech. It measures how natural and intelligible the speech sounds to a listener on a scale from 1 to 5, with 5 being the highest score. MOS is calculated by averaging the scores assigned by multiple human listeners who have evaluated the same set of audio samples.

For the evaluation, we have 8 different sources to be considered: the ground truth, baseline FastSpeech2, three SALTTS-parallel models and three SALTTS-cascade models. Ten audio samples were randomly sourced from each of these systems for human evaluation. These samples were then presented in a random order to 46 proficient English language speakers who were pursuing master's level education. The participants were asked to rate the audio samples on a scale of 1 to 5 based on the naturalness and intelligibility of the audios. Two separate sets of audio samples were created, each containing 5 samples from each source, and each set was evaluated by different set of 23 people. As a result, each human evaluator rated 40 audio samples sourced from 8 different systems.

MCD score as an evaluation metric is known to have its limitations. It may not be sensitive to some types of artifacts that affect speech quality, such as jitter, shimmer, and reverberation. MCD will not be able to capture the intelligibility of the speech as previously it has been observed that models having better MCD scores produce less intelligible speech \cite{tts_billion}.

The results for the experiments are shown in Table \ref{tab:Baseline Comparison}. Observing the objective measures (MCD and $F_0$ RMSE), all the models result in a similar performance compared to the baseline. However, the baseline FastSpeech2 has the least MCD score. All the proposed models outperforms the the baseline system in the log-$F_{0}$ RMSE score, although the difference is marginal. When it comes to the MOS score which is a subjective metric, the baseline FastSpeech2 performs relatively 3.8\% better compared to SALTTS-cascade (with data2vec-aqc), 3\% better compared to SALTTS-cascade (with ccc-wav2vec 2.0), and 1.9\%  better as compared to SALTTS-cascade (with HuBERT). However, all the proposed SALTTS-parallel models are better than the baseline. HuBERT (parallel) gave the best MOS score of 3.95 ± 0.18, which is relatively 8.2\% better than the baseline approach. The ccc-wav2vec 2.0 (parallel) and data2vec-aqc (parallel) showed a relative improvement of 6.02\% and 5.4\% compared to the FastSpeech2 baseline respectively. For both SALTTS-parallel and SALTTS-cascade models, the best SSL models turned out to be HuBERT followed by ccc-wav2vec 2.0 and data2vec-aqc.

We hypothesize that when using SALTTS-cascade models, the gradients responsible for the mel-spectrogram loss undergo a longer path before ultimately reaching the variance adapter and lower encoder layers, which could negatively impact performance. Although we attempted to enhance the model's performance by incorporating a residual connection from the Multi-layer Projector network to the mel-spectrogram decoder, the results did not show a significant improvement.

In contrast, for SALTTS-parallel models, the gradients responsible for mel-spectrogram loss immediately affect variance adapter and lower layer encoder. Furthermore, in a multi-task learning setup that includes the SSL embedding loss, the embedding after the Variance adapter tries to get richer, indirectly leading to better output from the mel-spectrogram decoder.

\vspace{-0.5em}
\section{Conclusion and Future work}
\vspace{-0.5em}

The present study proposes a novel approach called SALTTS (Self-supervised representations for Auxiliary Loss in TTS) to improve the quality of synthetic speech by incorporating more comprehensive speech-related information into TTS systems. Two variants of the proposed method are investigated: SALTTS-parallel and SALTTS-cascade. The results indicate that all SALTTS-parallel models outperform the baseline FastSpeech2 model, regardless of the SSL algorithm used. Conversely, none of the SALTTS-cascade models were able to surpass the baseline FastSpeech2 model. The SSL models explored in this paper are HuBERT, ccc-wav2vec2.0, and data2vec-aqc. The results showed that SALTTS-parallel, when used with HuBERT, provided the best performance with an 8.2\% relative improvement in MOS score over the baseline method FastSpeech2.

The main focus of this study is to investigate the impact of architectural modifications on the performance of TTS models. However, the study also highlights the potential for future research to analyze the effects of using other SSL models, such as WavLM, on the performance of TTS models. This avenue for further research could enhance our understanding of the complex relationships between TTS architectures and SSL methods.

\vspace{-0.5em}
\section{Acknowledgments}
\vspace{-0.5em}

Our sincere thanks to Anusha Prakash for her help in setting up the evaluation portal. We are thankful to Mudit Batra for his valuable inputs and support in the writing of this paper. We would like to thank the Ministry of Electronics and Information Technology (MeitY), Government of India, for providing us with the compute resources as a part of the ”Bhashini” project.



\vfill\pagebreak
\balance
\bibliographystyle{IEEEtran}
\bibliography{mybib}

\end{document}